\documentclass[11pt]{article}

\usepackage[margin=1in]{geometry} \usepackage{amsmath,amssymb,amsthm} \usepackage{booktabs} \usepackage{graphicx} \usepackage{tikz} \usetikzlibrary{arrows.meta,positioning} \usepackage[numbers,sort&compress]{natbib} \usepackage[colorlinks=true,linkcolor=blue,citecolor=blue,urlcolor=blue]{hyperref} \usepackage{microtype} \usepackage{times}

   \newcommand{\smin}{S_{\min}}

\title{Does Target Alignment Mean Target Recovery?\\ {\large An Evidence-Ladder Study of Adversarial Claims on Contrastive Encoders}}

\author{Tao Yang\\ Singapore University of Technology and Design\\ \texttt{tao\_yang@mymail.sutd.edu.sg} \and Jianying Zhou\\ Singapore University of Technology and Design\\ \texttt{jianying\_zhou@sutd.edu.sg}}

\date{October 2026}

\begin{document}

\maketitle

\begin{abstract}
Adversarial attacks on vision--language models optimize an image toward a text target, then cite the attacked model's similarity score as evidence of success. We ask whether that score---victim-space target alignment (VTS)---predicts recovery of the target by an independent model. We first build a measurement instrument: supervised judges outside the attacked geometry, real-target blend controls, shuffled-target negatives, and a reference level derived from a 50\% target-image blend. Two preregistered studies then compare six contrastive encoders under a matched attack at three perturbation budgets. Robustly trained encoders (FARE, TeCoA, PMG, TRADES) transfer substantially more independent evidence than vanilla CLIP or SigLIP; all eight contrasts reject at the bootstrap floor. However, no cell reaches the blend-derived reference level. The three best cells fall within its replication band, leaving practical recovery undecided. Within robust encoders, per-sample alignment gain correlates with evidence gain ($\rho=0.24$--$0.51$); within vanilla CLIP the correlation is consistent with zero. Across encoders we find no monotone alignment--evidence relation. VTS is therefore informative only within a fixed robust encoder, and we provide a reporting protocol in its place.

\end{abstract}

\section{Introduction} \label{sec:intro}

Optimizing an image to increase its similarity to a text target in a contrastive model's embedding space is a standard step in adversarial attacks on vision--language systems \citep{zhang2025anyattack,huang2025xtransfer,fang2025cpgc}. After optimization, the attacker reports the resulting score as evidence that the attack succeeded in inducing the target concept. The practice is widespread because the quantity is cheap to compute, directly optimized, and available to any white-box attack. But it is computed by the same encoder being optimized, so it rises under attack regardless of whether any other model can recognize the intended target. This circularity is not a hypothetical concern: prior work has shown that adversarial images can score high on victim-side alignment while failing entirely under independent evaluation. \citet{zhang2024illusions} found that adversarial images yield 0\% target recovery under an independent classifier while images generated from their embeddings yield 64\%, separating input-level alignment from generative recovery.

Despite this known gap, victim-side alignment remains the default evidence in adversarial semantic claims on contrastive models. Attacks are compared by their alignment scores, defenses are evaluated by how much they reduce alignment, and ``semantic'' conclusions are drawn from a quantity that the attacker controls. What remains unresolved is a narrower and more actionable question: \emph{when does increased alignment in the attacked model predict independently measured recovery?}

We approach this as a measurement problem rather than an attack-design problem. Our instrument pairs supervised judges, architecturally outside the contrastive family, with graded real-target blend controls and shuffled-target negatives, anchored to a reference level set by the mean independent evidence of a 50\% target-image blend. Two preregistered studies evaluate vanilla CLIP \citep{radford2021clip}, four adversarially fine-tuned variants (FARE, TeCoA, PMG, TRADES), and SigLIP \citep{zhai2023siglip} (a second contrastive family with a sigmoid objective), under a matched fixed-target attack at $\epsilon \in \{4,16,64\}/255$, on held-out images with regenerated target mappings. A preregistered recipe comparison additionally tests whether stronger standard attacks (targeted PGD, C\&W) change any conclusion.

Three results emerge. First, robustly fine-tuned encoders transfer more independent target evidence than non-adversarially trained ones. All eight robust-versus-naive contrasts reject at the bootstrap floor. SigLIP, which differs from vanilla CLIP in objective, scale, and training recipe, stays at vanilla-like levels. Second, no cell's mean evidence reaches the blend reference level. The three nearest cells fall within the replication band of that level, so the question of practical recovery remains open rather than settled. Third, alignment gain predicts evidence gain within robust encoders ($\rho = 0.24$--$0.51$) but not within vanilla CLIP (consistent with zero under a post-hoc approximation), and we find no monotone alignment--evidence relation across encoders. The naive-versus-robust inversion is real but does not extend to SigLIP.

An evidence taxonomy (L0 target intent through L4 human grounding) frames what each type of claim requires; our results are scoped to machine-level evidence (L1 and L2). Our contributions are:

\begin{enumerate}
\item \textbf{A calibrated measurement instrument} for independent target recovery, combining supervised judges outside the contrastive family, graded real-target blend positive controls, shuffled-target negatives, and a pre-specified reference level derived from a 50\% target-image blend. All controls are preregistered and their outputs are independently validated.
\item \textbf{A cross-encoder boundary map}: the first systematic evidence that adversarially fine-tuned encoders transfer more independent target evidence than non-adversarially trained ones, quantified across four robust recipes and two contrastive families (six encoders total) at three perturbation budgets under preregistered contrasts.
\item \textbf{A two-level diagnosticity finding}: victim-space alignment is informative within robust encoders ($\rho = 0.24$--$0.51$), consistent with zero within vanilla CLIP, and we find no monotone relation across encoders. This separates the metric's within-encoder use from its (unsupported) cross-encoder use.
\item \textbf{A reporting protocol} for adversarial semantic claims, specifying the evidence required at each level and the calibration steps that prior transfer-evaluation approaches lack.
\end{enumerate}

\section{Related Work} \label{sec:related}

\paragraph{Targeted and universal attacks on contrastive models.} Universal adversarial triggers originate in text classification \citep{wallace2019universal}; discrete gradient-based suffix search was systematized for LLM jailbreaks by GCG \citep{zou2023gcg}. Recent attacks on CLIP-scale models include targeted generators \citep{zhang2025anyattack} and universal perturbations \citep{fang2025cpgc,huang2025xtransfer}. Several evaluate on models beyond the surrogate (AnyAttack transfers to unseen target VLMs; \citet{springer2021robust} study robust-source targeted transfer), but their endpoint is attack or transfer success, not target-specific evidence calibrated against positive and negative controls.

\paragraph{Human evaluation and unrestricted attacks.} \citet{bhattad2020unrestricted} craft unrestricted semantic-manipulation attacks that preserve source-class content. \citet{elsayed2018fooling} show that adversarial perturbations bias time-limited human choices, reporting 57--89\% targeted transfer to models outside the optimization. \citet{veerabadran2023subtle} demonstrate that bounded perturbations bias human category judgments toward the chosen class. \citet{zhang2026concept} generate attacks whose original concept remains identifiable. These works establish that human evaluation of adversarial content has precedent; our contribution is the calibration protocol (graded controls, shuffle correction, a frozen reference level) applied across contrastive encoders and budgets, which none of them provide.

\paragraph{Robust features and gradient alignment.} Adversarial examples exploit predictive but non-robust features \citep{ilyas2019features}; robust classifiers carry perceptually aligned gradients \citep{santurkar2019synthesis,kaur2019pag}, which can be promoted without adversarial training while remaining associated with robustness \citep{ganz2023pag}. Off-manifold robustness helps explain this property \citep{srinivas2023offmanifold}. Gradient alignment in CLIP-family encoders has been characterized \citep{ganz2024clipag}; on-manifold perturbations relate robustness to generalization \citep{stutz2019disentangling,li2024manifold}; the joint space has a modality gap \citep{liang2022modality,levi2025doubleellipsoid}. None of this work conditions independent recoverability on victim robustness; the nearest results show that adversarially trained sources produce more transferable attacks \citep{awad2025defense} and that robustified CLIP induces better perceptual metrics \citep{croce2025perceptual}; neither endpoint is calibrated target evidence.

\paragraph{Evaluation validity.} Obfuscated gradients taught the field that evaluation under optimization requires adversarial scrutiny \citep{athalye2018obfuscated}; attack implementations and judge protocols remain sources of distortion \citep{cina2025attackbench,lorenz2021suitable,huang2025guidedbench,shen2025pandaguard,chen2025saferluckier,schwinn2026coinflip}, and human-grounded frameworks audit automated judges against perception \citep{scooter2025human}. Psychometric validity theory is being imported into ML evaluation \citep{bean2025measuring,salaudeen2025validity,freiesleben2026establishing}, covering both language and vision. No prior work addresses adversarial semantic claims with the calibrated protocol used here. We extend the evaluation-validity line from ``is the attack real?'' to ``does the semantic metric mean what it is used to mean?'' Universal-suffix work in the LLM domain independently supports non-semantic readings of trigger success \citep{bentov2025hijackers}.

\section{An Evidence Taxonomy for Adversarial Claims} \label{sec:ladder}

Adversarial ``semantic'' claims conflate several distinct questions. We separate them into levels: L0 target intent (WordNet or other ontology), L1 victim-space alignment (VTS), L2 independent-model recovery (ITE), L3 downstream propagation, and L4 human grounding. Passing one level does not imply passing another. Our results address L1 and L2; we do not claim L3 or L4 evidence.

\begin{table}[h] \centering \caption{Evidence levels and what each does not establish.} \label{tab:ladder} \small \begin{tabular}{llll} \toprule Level & Construct & Admissible evidence & Does not establish \\ \midrule L0 & Target intent & WordNet/ontology map & any change in the image \\ L1 & Victim alignment & frozen-target cosine/margin & independent recognition \\ L2 & Independent recovery & validated external judges & human perception \\ L3 & Downstream propagation & caption/VQA/retrieval endpoints & human perception \\ L4 & Human grounding & preregistered human study & universal meaning \\ \bottomrule \end{tabular} \end{table}

\textbf{Terminology.} VTS (victim-space target similarity) is cosine to the frozen target text embedding in the victim. ITE (independent target evidence) is the shuffle-corrected change in the target's logit margin on an independent judge. ITR (rank recovery, anchor in top-5) is corroborating secondary. Calling an L1 quantity ``semantic injection'' or ``semantic coherence'' without L2 or higher evidence is a category error under this framework; we apply the same rule to our own claims.

This taxonomy draws on validity-evidence thinking from measurement theory \citep{cronbach1955construct,messick1995validity}, adapted for adversarial evaluation. Concurrent work imports the same theory for ML evaluation more broadly \citep{bean2025measuring,salaudeen2025validity,freiesleben2026establishing}, covering both language and vision; ours applies it specifically to adversarial semantic claims on vision--language models, with the calibrated protocol (blend positive controls, shuffled negatives, a frozen reference level) that prior frameworks do not provide. The mapping to Messick's validity aspects (L1 as structural, L2 as external/convergent, L3 as consequential, L4 as substantive-plus-content) is heuristic rather than established; it serves as a bridge, not a derivation.

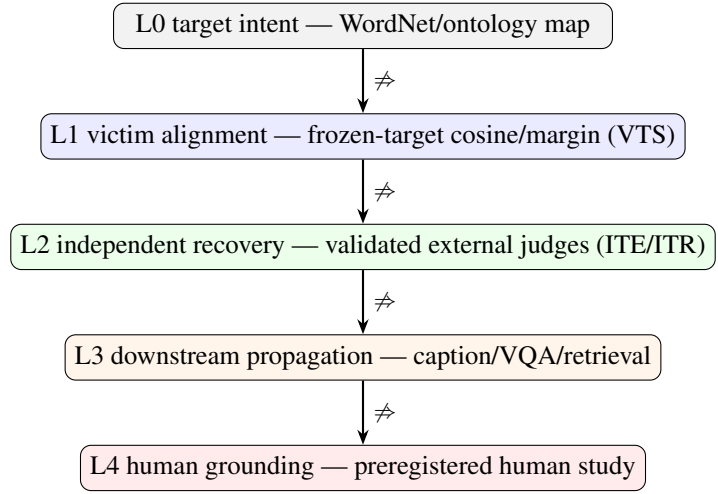
\begin{figure}[t] \centering \begin{tikzpicture}[node distance=0.85cm, lvl/.style={draw, rounded corners, minimum width=6.6cm, minimum height=0.6cm, align=center, font=\small}] \node[lvl, fill=gray!10] (l0) {L0 target intent --- WordNet/ontology map}; \node[lvl, below=of l0, fill=blue!8] (l1) {L1 victim alignment --- frozen-target cosine/margin (VTS)}; \node[lvl, below=of l1, fill=green!8] (l2) {L2 independent recovery --- validated external judges (ITE/ITR)}; \node[lvl, below=of l2, fill=orange!8](l3) {L3 downstream propagation --- caption/VQA/retrieval}; \node[lvl, below=of l3, fill=red!8] (l4) {L4 human grounding --- preregistered human study}; \foreach \i/\j in {l0/l1, l1/l2, l2/l3, l3/l4}{ \draw[-{Stealth[length=2mm]}, thick] (\i) -- node[right, font=\scriptsize] {$\not\Rightarrow$} (\j); } \end{tikzpicture} \caption{Evidence levels. Adjacent levels do not imply one another.} \label{fig:ladder} \end{figure}

\section{Methods} \label{sec:methods}

\subsection{Study A: instrument validation} \label{sec:methods-a}

\textbf{Judges.} The primary judge is ResNet-50 (ImageNet-1k, supervised), chosen for architectural and training disjointness from the ViT-B/32 victims. The secondary judge, ViT-B/16 (supervised), is reported in the artifacts; it shares the ViT family with the victims, so it corroborates direction and ordering but not scale. Both judges operate on the same 1000-class ImageNet label space, which is the space our evaluation classes map into.

\textbf{Controls.} PC1 (judge competence): clean target-class exemplars establish per-class coverage. PC2 (graded positive control): source images blended with a \emph{real} target-class image at doses $\alpha\in\{0.25,0.5,0.75,1.0\}$; the blend constructor is never used as a judge. NC (negatives): clean sources and shuffled targets. A null attack result is interpretable only when these pass on the same endpoint and judge.

\textbf{Statistic.} For judge logits $z_c$, the target logit margin is $m_a = z_a - \operatorname{logmeanexp}_{c\neq a} z_c$. The intended change is $D_a = m_a(x_{\mathrm{adv}}) - m_a(x)$; the target-specific statistic is $S_a = D_a - \mathbb{E}_{a'}[m_{a'}(x_{\mathrm{adv}}) - m_{a'}(x)]$, where the expectation runs over 16 shuffled targets $a'$ per image (drawn without replacement from the 1000 judge labels, excluding $a$ and the true class $y$). The image is never regenerated under shuffling. We denote $S_a$ the ITE.

\textbf{Reference level.} $\smin$ is anchored at the $\alpha^{*}=0.50$ blend dose (half of each source pixel replaced by a real target image) and was frozen, with hashes, before Study B (Table~\ref{tab:study-a}). A cell is \emph{L2-positive at the anchor} iff its mean ITE $\ge \smin$ and its class-cluster bootstrap 95\% CI lower bound exceeds zero. We treat $\smin$ as a pre-specified, reproducible reference level, not a demonstrated utility threshold; utility calibration would require an external task payoff.

\subsection{Study B: confirmatory core} \label{sec:methods-b}

\textbf{Victims.} Vanilla CLIP, FARE, and TeCoA (all ViT-B/32). TeCoA was trained at $\epsilon_{\mathrm{train}}{=}1/255$ \citep{mao2023tecoa}; our FARE checkpoint is the $\epsilon_{\mathrm{train}}{=}4/255$ release (FARE$^4$; hash-pinned in the frozen manifest) \citep{schlarmann2024fare}. We evaluate at $\epsilon\in\{4,16,64\}/255$, which spans $1$--$16\times$ the FARE and $4$--$64\times$ the TeCoA training budgets.

\textbf{Attack.} The attack minimizes a two-term objective. The first term increases target alignment: $-0.4\,\langle \hat f_I(x_i{+}\delta_i), \hat e_{a_i}\rangle$, where $\hat f_I$ is the $\ell_2$-normalized image embedding and $\hat e_{a_i}$ the frozen target text embedding (both cosines). The second term decreases source-class alignment: $\lambda_i \langle \hat f_I(x_i{+}\delta_i), \hat e_{y_i}\rangle$, where $\hat e_{y_i}$ is the true-class embedding. The weight $\lambda_i$ is 0.5 while the victim's 100-class head still predicts $y_i$, otherwise 0.1, re-evaluated after each optimization step. Adam optimizer, learning rate 0.01, 20 steps, $\ell_\infty$ projection each step.

\textbf{Data.} Targets are regenerated per mapping seed $\{101,202,303\}$. Each seed draws 100 images from the victim-clean-correct intersection of the confirmatory rows $[2000,5000)$ of a fixed permutation of the ImageNet-100 validation set (exploratory work used rows $[0,2000)$). Each cell thus comprises 300 paired image-mapping observations across three seeds; the seeds share the row pool, so unique source images number approximately 285 per cell.

\textbf{Power and clustering.} Cell size was set from pilot estimates of the paired statistic's standard deviation (1.7--4.1 ITE units on R50). All confirmatory mean contrasts and cell intervals use the class-cluster bootstrap. The within-cell intra-class correlation of ITE over classes is small to moderate (ICC(1,1), unequal-size-corrected: $-0.03$ to $0.21$). The correlation and equivalence analyses in \S\ref{sec:results-v2} use independent-sample Fisher/TOST approximations, reported as post-hoc sensitivity.

\textbf{Freeze.} Design freeze (grid, $\smin$, splits, hashes) and code freeze (runner, lockout tests) are separate tagged commits. Outputs are append-only per seed.

\subsection{Study C: additional encoders} \label{sec:methods-c}

A second, separately frozen design (same rows, seeds, statistic, judge, rule) widens the victim set to six. The robust recipes PMG and TRADES are local adversarially fine-tuned checkpoints, CLIP ViT-B/32-compatible, with recipes partially following \citet{wang2024pmgaft} and a TRADES-style objective \citep{zhang2019trades}; all checkpoints are hash-pinned in the frozen manifests. SigLIP (\texttt{siglip-base-patch16-224}, sigmoid objective) uses its own text tower and normalization. Both towers use the same prompt in their own tokenizers; VTS is computed within-tower.

Predeclared hypotheses. H\_C1: PMG and TRADES exceed vanilla at $\epsilon\in\{16,64\}$ (Holm-4, paired). H\_C2: SigLIP stays vanilla-like (descriptive; no calibrated equivalence margin exists for a new family). H\_C3: victim-mean VTS anti-ranks ITE across the six encoders (direction-only).

Study C shares its confirmatory rows, mapping seeds, and anchor/shuffle constructions with Study B; the cohorts differ only by the intersection rule. The two Holm families are corrected separately. No pooled cross-study confirmatory analysis is performed; the post-hoc prediction check in \S\ref{sec:discussion} pools records as a declared exploratory exception. The dataset axis (fine-grained, texture) is deferred: those classes are absent from the ImageNet-1k judge's label space.

\textbf{Analysis.} Cell-level ITE with class-cluster bootstrap. Four one-sided robust-versus-vanilla contrasts on paired ITE differences under step-down Holm ($\alpha=0.05$, six-contrast family). The family also contains two FARE-versus-TeCoA contrasts; specified one-sided, they were guaranteed-null under the observed ordering, which makes the correction conservative for the four boundary tests. Within-cell Spearman $\rho(\Delta\mathrm{VTS}, S_a)$ is the preregistered primary form; between-victim means use post-attack VTS. All cells are reported; no re-derivation under alternative rules.

\section{Results} \label{sec:results}

\subsection{Instrument validation} \label{sec:results-instrument}

\begin{figure}[t] \centering \includegraphics[width=0.55\linewidth]{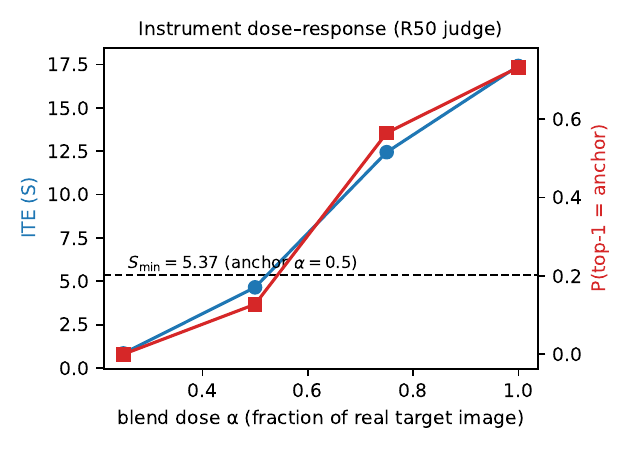} \caption{Instrument dose--response on the R50 judge: ITE (left axis) and top-1 recovery (right axis) vs.\ blend dose. The reference level equals the ITE at the $\alpha{=}0.5$ dose.} \label{fig:dose} \end{figure}

Study A passed all controls. The primary judge recovers real target evidence monotonically in blend dose ($\rho=1.0$, Figure~\ref{fig:dose}). The anchor-dose ITE is $5.71$ [95\% CI $5.12$, $6.32$] on held-out data; the reference level is $\smin=5.3711$ (Table~\ref{tab:study-a}). Negative controls pass on the same endpoint. This establishes that the instrument measures machine-grounded target evidence under the stated protocol; nothing here addresses human-perceived semantics.

\begin{table}[t]
\centering
\caption{Study A: validation of the L2 independent-evidence instrument
(held-out split, $n{=}250$). The pre-specified reference level is anchored at the
$\alpha^{*}{=}0.50$ blend dose (half of each source image's pixels replaced by
a real target-class image) and frozen before Study B.}
\label{tab:study-a}
\small
\begin{tabular}{lr}
\toprule
Quantity & Value \\
\midrule
Primary judge & ResNet-50 (ImageNet-1k, supervised) \\
Gate A (PC1/PC2/negative controls) & \textbf{PASS} \\
Held-out ITE at the anchor dose & $5.71$ [95\% CI $5.12$, $6.32$] \\
Held-out ITR top-5 increase & $0.324$ [95\% CI $0.268$, $0.384$] \\
Dose--response (Spearman) & $\rho = 1.00$ (monotone) \\
Frozen margin $S_{\min}$ & $5.3711$ \\
\bottomrule
\end{tabular}
\end{table}

\subsection{Robustness--evidence contrasts} \label{sec:results-boundary}

Independent target evidence rises with budget for every encoder (Table~\ref{tab:holm}; per-cell values in Table~\ref{tab:study-b}). At $\epsilon{=}16$ and $64$ the ordering is TeCoA $>$ FARE $>$ vanilla; at $\epsilon{=}4$ FARE and TeCoA do not separate. All four preregistered contrasts reject at the bootstrap floor ($p = 1/B = 0.0005$, step-down Holm), with mean differences from $+1.83$ to $+4.45$ ITE units over 300 paired images.

\begin{table}[t]
\centering
\caption{Study B boundary contrasts (one-sided robust $>$ base on paired
ITE differences, $n{=}300$ pairs each; class-cluster bootstrap;
step-down Holm over the 6-contrast family, $\alpha{=}0.05$).}
\label{tab:holm}
\footnotesize
\begin{tabular}{lrrr}
\toprule
Contrast & Pairs & Mean difference & Holm reject \\
\midrule
FARE-vanilla @ $\epsilon{=}$16 & 300 & +1.83 & \textbf{yes} \\
FARE-vanilla @ $\epsilon{=}$64 & 300 & +2.60 & \textbf{yes} \\
TeCoA-vanilla @ $\epsilon{=}$16 & 300 & +2.55 & \textbf{yes} \\
TeCoA-vanilla @ $\epsilon{=}$64 & 300 & +4.45 & \textbf{yes} \\
FARE-TeCoA @ $\epsilon{=}$16 & 300 & -0.72 & no \\
FARE-TeCoA @ $\epsilon{=}$64 & 300 & -1.85 & no \\
\bottomrule
\end{tabular}
\end{table}

No cell reaches the reference level. Table~\ref{tab:study-b} reports per-cell values. At $\epsilon{=}4$, all six encoders cluster near zero ITE ($0.08$--$0.39$), with overlapping CIs. At $\epsilon{=}16$, the robust recipes separate from vanilla by $+2.3$ to $+2.6$ ITE units. At $\epsilon{=}64$, the gap widens to $+3.4$ to $+4.5$ units for TeCoA, $+3.4$ for FARE, and $+4.2$--$+4.9$ in Study C for PMG and TRADES. The best cell (TeCoA, $\epsilon{=}64$) has mean ITE $5.244$ with CI lower bound $4.74$: the CI condition passes, the magnitude condition fails by 2.4\%. An independent replication of the anchor dose on a different cohort gave $4.65$ (approximately 15\% calibration sensitivity). Table~\ref{tab:anchor-sens} gives each cell's verdict under both anchors: exactly three cells (B:TeCoA@64, C:TeCoA@64, C:PMG@64) sit inside the band and flip; all others are negative under both. No cell passes at the Study-A held-out anchor ($5.71$) either, so the flip set is stable across the full $[4.65, 5.71]$ range of anchor estimates.

\begin{table}[t]
\centering
\caption{Anchor-sensitivity of the L2-positive-at-the-anchor verdict (all 27 confirmatory cells). A cell counts iff mean ITE $\ge$ anchor AND CI lower $>0$ (CI condition holds in every cell). Frozen anchor $5.371$; replicated anchor dose $4.65$. Exactly three cells sit inside the band and flip; every other cell is negative under both.}
\label{tab:anchor-sens}
\small
\begin{tabular}{lrrcc}
\toprule
Cell (study, victim, $\epsilon$) & mean & CI low & $\ge\!5.371$ & $\ge\!4.65$ \\
\midrule
B: TeCoA@64 & 5.24 & 4.74 & no & yes \textbf{flip} \\
C: TeCoA@64 & 4.98 & 4.35 & no & yes \textbf{flip} \\
C: PMG@64 & 4.90 & 4.42 & no & yes \textbf{flip} \\
C: TRADES@64 & 4.30 & 3.85 & no & no  \\
\midrule
\multicolumn{5}{l}{All remaining 23 cells: mean ITE $<4.0$, negative under both anchors} \\
\bottomrule
\end{tabular}
\end{table}

\subsection{Recipe comparison} \label{sec:results-frontier}

The preregistered recipe could in principle sit far below the achievable optimization frontier, reducing the boundary claim to a statement about one optimizer. We ran targeted PGD and C\&W-style attacks (100 iterations each) alongside the preregistered recipe on exploration-only data, under the identical cohort, anchors, shuffles, and judge. No variant crosses $\smin$ in any cell; the preregistered recipe itself achieves the highest ITE in five of six victim-budget combinations. PGD-100 attains higher victim-side alignment (VTS up to 0.63 vs.\ 0.49) while transferring equal or less evidence. This is a within-optimization replication of the L1$\perp$L2 dissociation: pushing the victim's target score harder does not buy more external evidence (Table~\ref{tab:frontier}).

\subsection{Within-encoder and cross-encoder diagnosticity} \label{sec:results-v2}

Within every robust-encoder cell, per-sample alignment gain predicts evidence gain ($\rho = 0.24$--$0.51$; Fisher 95\% CIs exclude zero, e.g.\ TeCoA@16 $[0.42, 0.59]$; stable across mapping seeds). Within vanilla CLIP, the observed $\rho \in [-0.04, 0.00]$ carry Fisher 95\% CIs no wider than $[-0.15, 0.11]$. A post-hoc TOST against an exploratory small-association tolerance $|\rho| \ge 0.15$ (conventional bound; post-confirmatory, independent-sample approximation, clustering not accounted for) gives $p = 0.028/0.010/0.005$ at $\epsilon = 4/16/64$, consistent with but not proving a negligible association. The TOST covers vanilla only. Between victims in Study B, the mean-VTS ranking inverts the mean-ITE ranking ($\rho = -1.0$; $n{=}3$, descriptive): vanilla has the highest post-attack alignment and the lowest evidence. Pooling the nine cells gives $\rho \approx +0.08$; the two levels cancel.

\subsection{Study C: additional encoders} \label{sec:results-c}

A separately frozen design extends the grid to PMG, TRADES, and SigLIP on a six-encoder intersection cohort (Table~\ref{tab:study-c} in the appendix).

\textbf{H\_C1: robustness extension.} All four Holm-corrected contrasts reject at the bootstrap floor. PMG$-$vanilla: $+2.53$/$+4.24$; TRADES$-$vanilla: $+2.33$/$+3.64$ ITE units at $\epsilon{=}16/64$. The gap between robust and naive now replicates on four distinct recipes, whose $\epsilon{=}64$ cells cluster tightly (ITE $3.3$--$5.0$).

\textbf{H\_C2: family extension (descriptive).} SigLIP remains at vanilla-like levels at every budget (ITE $0.08$--$0.35$). SigLIP differs from the CLIP encoders in objective, scale, and training recipe; this comparison does not isolate the cause of the robust-naive gap. The four-recipe robust-versus-naive contrast is the strongest available, but it remains a checkpoint comparison rather than a matched training intervention.

\textbf{H\_C3: cross-encoder anti-ordering (preregistered failure).} The direction-only prediction that victim-mean VTS anti-ranks ITE does not survive at $n{=}6$ ($\rho = +0.03$, $p = 0.96$, computed at $\epsilon{=}64$ on the Study-C cohort): SigLIP has both the lowest VTS and the lowest ITE. The Study-B inversion is real but conditional. Across all six encoders, we find no monotone alignment--evidence relation in either direction. This strengthens, rather than weakens, the practical conclusion: cross-encoder VTS comparison fails in a way that includes inversion between regimes but is not reducible to it.

\begin{figure}[t] \centering \includegraphics[width=\linewidth]{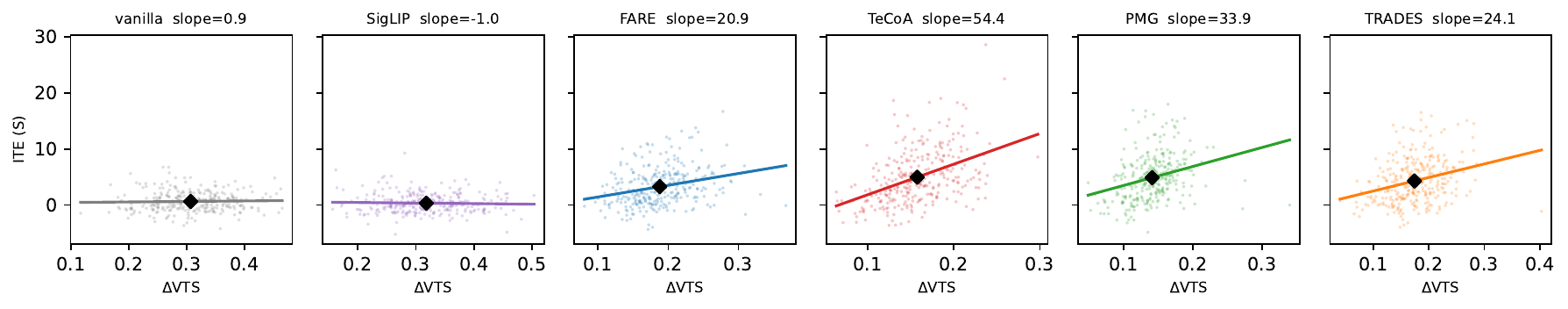} \caption{Within-encoder slopes at $\epsilon{=}64/255$ across six encoders (Study-C cohort, $n{=}300$ per panel). X-axis ranges vary; line angles are not directly comparable.} \label{fig:twolevel} \end{figure}

\subsection{Secondary observations} \label{sec:results-secondary}

ITR (anchor in top-5) corroborates the ordering: TeCoA@64 $0.293$, PMG@64 $0.263$, TRADES@64 $0.213$, FARE@64 $0.180$, vanilla@64 $0.013$, SigLIP@64 $0.013$. That nearly one in three test images of TeCoA@64 has the attack-chosen target class in the independent judge's top-5 confirms that the ITE signal is not a statistical artifact, even though the mean does not reach the reference level. The secondary ViT-B/16 judge corroborates direction and ordering at $\epsilon{=}16$ and $64$, with per-sample cross-judge Spearman $\rho = 0.55$--$0.59$ among robust victims. At $\epsilon{=}4$ it shows no robust-victim elevation; those cells are descriptive in the frozen design. Cross-judge logit scales are uncalibrated, so the ViT judge is not compared to the R50-derived reference level. Low vanilla cross-judge agreement ($\rho \approx 0.20$--$0.25$) reflects the near-zero signal range.

\begin{table}[t]
\centering
\caption{Study B (confirmatory): independent target evidence ITE$_{S_a}$
per cell, pooled over mapping seeds 101/202/303 ($n{=}300$/cell; R50 judge;
held-out ImageNet-100 rows). Rule: L2-positive at the anchor $\iff$ mean
$\ge S_{\min}=5.371$ \emph{and} class-cluster bootstrap 95\% CI lower bound
$>0$. ITR = anchor-in-top-5 rate. VTS = victim-space target similarity after
attack (L1 diagnostic).}
\label{tab:study-b}
\footnotesize
\begin{tabular}{llrrrrrr}
\toprule
Victim & $\epsilon$ & $n$ & ITE mean & ITE sd & CI$_{95}$ lower & ITR top-5 & VTS \\
\midrule
vanilla & 4/255 & 300 & 0.1653 & 0.6131 & 0.1015 & 0.0033 & 0.4298 \\
vanilla & 16/255 & 300 & 0.611 & 1.6177 & 0.4464 & 0.01 & 0.4784 \\
vanilla & 64/255 & 300 & 0.7935 & 1.9026 & 0.5988 & 0.0133 & 0.4879 \\
FARE & 4/255 & 300 & 0.3117 & 0.8091 & 0.2361 & 0.0067 & 0.2362 \\
FARE & 16/255 & 300 & 2.4412 & 2.7311 & 2.1131 & 0.08 & 0.3499 \\
FARE & 64/255 & 300 & 3.3974 & 3.2656 & 3.025 & 0.19 & 0.3842 \\
TeCoA & 4/255 & 300 & 0.2903 & 0.6925 & 0.2191 & 0.0033 & 0.1871 \\
TeCoA & 16/255 & 300 & 3.162 & 3.3292 & 2.781 & 0.1233 & 0.2778 \\
TeCoA & 64/255 & 300 & 5.2441 & 4.4289 & 4.7433 & 0.2933 & 0.3222 \\
\bottomrule
\end{tabular}

\smallskip
{\footnotesize $\dagger$ inconclusive-at-margin: CI condition passes, mean
within 0.5 of $S_{\min}$ (see Sensitivity). No cell is L2-positive
at the anchor.}
\end{table}

\begin{table}[t]
\centering
\caption{Two-level diagnosticity of victim-space alignment (confirmatory).
Within-cell: Spearman $\rho$ between per-sample $\Delta$VTS and ITE$_{S_a}$
(all permutation $p \le 0.0005$ except vanilla, ns). Between-victim:
victim-mean VTS vs victim-mean ITE. Pooled cell-level $\rho$ shown to exhibit
the Simpson cancellation that pooled correlations conceal.}
\label{tab:v2}
\footnotesize
\begin{tabular}{lrrr}
\toprule
Victim & $\rho$ ($\epsilon{=}4$) & $\rho$ ($\epsilon{=}16$) & $\rho$ ($\epsilon{=}64$) \\
\midrule
vanilla & -0.041 & -0.016 & -0.001 \\
FARE & +0.314 & +0.368 & +0.236 \\
TeCoA & +0.313 & +0.512 & +0.409 \\
\midrule
\multicolumn{4}{l}{Between-victim means (van/FARE/TeCoA):} \\
\multicolumn{4}{l}{VTS $0.465/0.323/0.262$ vs ITE $0.52/2.05/2.90$;} \\
\multicolumn{4}{l}{$\rho=-1.0$ ($n{=}3$, descriptive)} \\
Pooled cell-level (9 cells): $\rho = +0.083$ (cancellation) \\
\bottomrule
\end{tabular}
\end{table}

\begin{table}[t]
\centering
\caption{Study C per-cell results by budget (confirmatory; $n{=}300$/cell; R50 judge; cf.\ Holm-4 and H\_C2/H\_C3 in the main text). No cell is L2-positive at the anchor.}
\label{tab:study-c}
\footnotesize
\begin{tabular}{llrrrrrr}
\toprule
$\epsilon$ & Metric & vanilla & FARE & TeCoA & PMG & TRADES & SigLIP \\
\midrule
4/255 & ITE mean & 0.1961 & 0.3288 & 0.2786 & 0.3893 & 0.3748 & 0.0833 \\
 & CI$_{95}$ lower & 0.1288 & 0.2417 & 0.1987 & 0.3094 & 0.2954 & 0.0196 \\
 & ITR top-5 & 0.01 & 0.0033 & 0.0067 & 0.01 & 0.0033 & 0.0 \\
 & VTS & 0.4325 & 0.2342 & 0.1846 & 0.2032 & 0.2075 & 0.2367 \\
\midrule
16/255 & ITE mean & 0.5428 & 2.3106 & 2.9692 & 3.0684 & 2.8698 & 0.2907 \\
 & CI$_{95}$ lower & 0.3955 & 1.9795 & 2.5086 & 2.6842 & 2.4941 & 0.1533 \\
 & ITR top-5 & 0.02 & 0.0833 & 0.14 & 0.13 & 0.1333 & 0.01 \\
 & VTS & 0.4811 & 0.3478 & 0.2757 & 0.2845 & 0.2873 & 0.2788 \\
\midrule
64/255 & ITE mean & 0.6595 & 3.2891 & 4.9815 & 4.8988 & 4.302 & 0.3493 \\
 & CI$_{95}$ lower & 0.497 & 2.8931 & 4.3486 & 4.4232 & 3.8512 & 0.1969 \\
 & ITR top-5 & 0.02 & 0.18 & 0.2667 & 0.2633 & 0.2133 & 0.0133 \\
 & VTS & 0.4913 & 0.3827 & 0.3207 & 0.3213 & 0.3182 & 0.2801 \\
\bottomrule
\end{tabular}
\end{table}

\begin{table}[t]
\centering
\caption{Recipe-frontier check (exploratory split; identical cohort, anchors, and judge to the calibration pilot). ITE under the preregistered recipe (frozen-20) vs standard stronger optimizers at 100 iterations. No variant crosses $S_{\min}=5.371$ in any cell; the preregistered recipe attains the best ITE in five of six victim--budget combinations. PGD-100 attains higher victim-side alignment (VTS) than the recipe while transferring less evidence.}
\label{tab:frontier}
\footnotesize
\begin{tabular}{llrrrrrr}
\toprule
Victim & $\epsilon$ & \multicolumn{3}{c}{ITE mean} & \multicolumn{3}{c}{VTS} \\
 & & frozen & PGD-100 & C\&W-100 & frozen & PGD-100 & C\&W-100 \\
\midrule
vanilla & 16/255 & 0.6571 & 0.3098 & 0.3626 & 0.4789 & 0.5723 & 0.4898 \\
vanilla & 64/255 & 0.7479 & 0.6111 & 0.5009 & 0.4883 & 0.63 & 0.5029 \\
FARE & 16/255 & 2.569 & 2.4876 & 2.0176 & 0.3483 & 0.4254 & 0.3705 \\
FARE & 64/255 & 3.3824 & 2.991 & 1.5034 & 0.383 & 0.5423 & 0.4411 \\
TeCoA & 16/255 & 3.1555 & 3.4647 & 3.2887 & 0.2762 & 0.3296 & 0.3 \\
TeCoA & 64/255 & 5.0246 & 4.5935 & 3.0152 & 0.3198 & 0.4508 & 0.3839 \\
\bottomrule
\end{tabular}
\end{table}

\section{Discussion} \label{sec:discussion}

\paragraph{What VTS measures and does not measure.} Within a robustly fine-tuned encoder, VTS tracks independent evidence per sample. Within vanilla CLIP, the monotone association is consistent with zero under a post-hoc approximation. Across encoders, the naive-versus-robust inversion does not extend to SigLIP. These three facts together limit VTS to a within-encoder diagnostic for robust checkpoints. To quantify the cross-encoder prediction question, we ran a post-hoc sensitivity check (not preregistered; class-grouped 5-fold cross-validation over 8,100 pooled records): adding $\Delta$VTS to an additive encoder-plus-budget baseline changes top-5-recovery AUC from 0.829 to 0.839 and $R^2$ from 0.236 to 0.288. These are small positive increments under this specific baseline, split, and set of endpoints. The folds hold out source classes, not encoders, so the result applies to seen encoders and does not establish or rule out incremental value in other settings.

\paragraph{The robustness--evidence gap.} Two findings sit side by side. Robust recipes show higher mean ITE than naive ones: large effects, monotone in budget, Holm-corrected. Within robust cells, alignment gain correlates with evidence gain; the correlation itself is not monotone in budget (FARE peaks at $\epsilon{=}16$). Neither finding is a training-causal claim, because the encoders are released checkpoints rather than matched interventions. No cell's mean reaches the reference level, even at $\epsilon{=}64/255$, an unusually large budget. The evaluation budgets exceed the reported TeCoA training budget; FARE's exact training budget is unresolved. The contrasts are budget-matched, so the robust-naive gap is not explained by budget differences, but recipe attribution and interpretation relative to training budgets require verified provenance.

\paragraph{Near-threshold cells.} The three best cells sit between the two anchor calibrations ($5.371$ frozen, $4.65$ replicated). We report them as inconclusive rather than resolved in either direction. The anchor's calibration uncertainty (approximately 15\% between draws) is comparable to the gap between these cells and the reference level, which is exactly the situation where a binary verdict would be misleading.

\paragraph{A reporting protocol for adversarial semantic claims.} No existing standard covers what an adversarial ``semantic'' claim needs to report. In place of raw VTS, we recommend: (i) scope alignment claims to the evaluated encoder; (ii) report at least one validated independent endpoint, with graded real-target positive controls, shuffled-target negatives, and a judge competence check; (iii) state the optimization recipe and its distance to standard stronger attacks; (iv) for robust encoders, evaluate at and just beyond the training budget. These four items distinguish the calibration protocol from prior transfer-evaluation approaches.

\paragraph{Checkpoint provenance.} Our FARE checkpoint is labeled $\epsilon_{\mathrm{train}}{=}4/255$ based on the artifact-repository README, but the RobustVLM release lists multiple FARE variants for ViT-B/32 ($\epsilon_{\mathrm{train}}{=}1$ and $4/255$), and we have not independently verified which upstream checkpoint produced our local file. The checkpoint's SHA-256 hash matches the frozen manifest, so the experiments are internally consistent, but the training-budget attribution and the regime-mismatch interpretation depend on resolving this provenance question. TeCoA's $1/255$ budget is confirmed from its paper.

\paragraph{What the protocol adds.} The standard evaluation in this space reports attack success rate or victim-side alignment. Our instrument adds three things: graded real-target blend controls that establish the judge's sensitivity to actual target evidence; shuffled-target negatives that correct for non-specific logit-mass movement; and a pre-specified reference level that anchors the verdict to a measurable quantity rather than a threshold chosen post hoc. Any of these can be adopted independently of the others.

\paragraph{Mechanism.} The observed pattern is consistent with robust-features and off-manifold accounts \citep{ilyas2019features,santurkar2019synthesis,srinivas2023offmanifold,stutz2019disentangling,awad2025defense}, but no mechanism quantity was measured confirmatorily. Identifying the causal pathway from adversarial training to independent evidence transfer would require matched training interventions or a continuous robustness axis, which are outside the scope of this paper.

\paragraph{Limitations.} Six encoders (five ViT-B/32-family recipes plus SigLIP), one dataset (ImageNet-100), one attack objective (fixed-target), three budgets, one preregistered recipe. The preregistered cross-encoder anti-ordering hypothesis failed ($\rho\approx 0$); the between-encoder claim is therefore stated conditionally. Fine-grained and texture datasets were not run (classes outside the judge's label space). Encoders are released checkpoints, not matched training interventions; FARE's training budget is unresolved. Claims are capped at L2 under supervised ImageNet-1k judges (no L3 or L4 endpoint; the secondary judge shares the ViT family). The reference level inherits approximately 15\% inter-draw uncertainty. No mechanism quantity was measured confirmatorily. The adaptive-objective arm was dropped by its own preregistered gate.

\paragraph{Future work.} Several extensions follow from these findings. Matched training interventions that vary adversarial-training strength along a continuum would isolate whether robustness itself causes the alignment--evidence gap or whether the gap tracks a correlated property of the released checkpoints. Extending the instrument to additional judge paradigms (self-supervised probes, zero-shot captioning models) would test whether the L2 boundary is judge-specific or a property of the encoders themselves. Downstream propagation (L3) remains untested: if the classifier-level evidence we measure propagates to generative or retrieval endpoints, the sub-threshold finding may understate the practical risk. The reporting protocol could be developed into a community benchmark covering additional attack families, encoder architectures, and perturbation norms. A human-grounded study (L4) would determine whether the blend-derived reference level corresponds to any perceptual threshold.

\bibliography{references}
\bibliographystyle{plainnat}

\end{document}